\documentclass[sigconf]{acmart}
\usepackage{booktabs}
\usepackage{caption}
\usepackage[table,xcdraw]{xcolor}
\renewcommand\footnotetextcopyrightpermission[1]{} 
\AtBeginDocument{%
  }

\copyrightyear{2026}
\acmYear{2026}
\setcopyright{cc}
\setcctype{by}
\acmConference[GLSVLSI '26]{Great Lakes Symposium on VLSI 2026}{June 22--24, 2026}{Canandaigua, NY, USA}
\acmBooktitle{Great Lakes Symposium on VLSI 2026 (GLSVLSI '26), June 22--24, 2026, Canandaigua, NY, USA}
\acmDOI{10.1145/3787109.3816372}
\acmISBN{979-8-4007-2431-2/2026/06}

\begin{document}

\title{FAR: Function-preserving Attention Replacement for IMC-friendly Inference}


\author{Yuxin Ren}
\affiliation{%
  \institution{University of Arizona}
  \city{Tucson}
  \country{USA}}
\email{yuxinr@arizona.edu}

\author{Maxwell D Collins}
\affiliation{%
  \institution{TetraMem, Inc.}
  \city{San Jose}
  \country{USA}}
\email{maxwell.collins@tetramem.com}

\author{Miao Hu}
\affiliation{%
  \institution{TetraMem, Inc.}
  \city{San Jose}
  \country{USA}}
\email{miao.hu@tetramem.com}

\author{Huanrui Yang}
\affiliation{%
  \institution{University of Arizona}
  \city{Tucson}
  \country{USA}}
\email{huanruiyang@arizona.edu}



\begin{abstract}
While transformers dominate modern vision and language models, their attention mechanism remains poorly suited for in-memory computing (IMC) devices due to intensive activation-to-activation multiplications and non-local memory access, leading to substantial latency and bandwidth overhead on ReRAM-based accelerators. 
To address this mismatch, we propose FAR, a Function-preserving Attention Replacement framework that substitutes all attention in pretrained DeiTs with sequential modules inherently compatible with IMC dataflows. 
Specifically, FAR replaces self-attention with a multi-head bidirectional LSTM architecture via block-wise distillation to retain functional equivalence while enabling linear-time computation and localized weight reuse. 
We further incorporate structured pruning on FAR models, enabling flexible adaptation to resource-constrained IMC arrays while maintaining functional fidelity.
Evaluations on the DeiT family demonstrate that FAR maintains comparable accuracy to the original attention-based models on ImageNet and multiple downstream tasks with reduced parameters and latency. 
Further analysis shows that FAR preserves the semantic token relationships learned by attention while improving computational efficiency, highlighting its potential for energy-efficient transformer inference on IMC-based edge accelerators.
\end{abstract}





\maketitle
\pagestyle{plain}

\section{Introduction}
\label{sec:intro}

\begin{figure}[t]
\centering
\includegraphics[width=.75\linewidth]{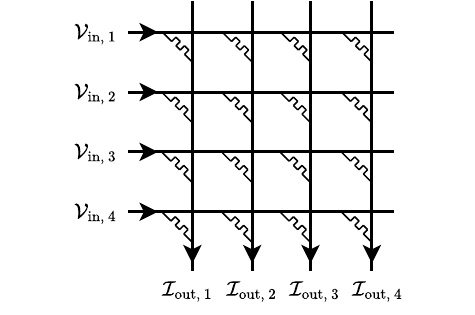}
\caption{IMC crossbar illustration}
\Description{IMC crossbar illustration}
\label{fig:crossbar}
\end{figure}

Transformer architectures have become the dominant backbone in multimodal, vision, and language tasks~\cite{vaswani2017attention, devlin2019bert, dosovitskiy2021vit, radford2021clip}. 
Their effectiveness largely due to self-attention, which enables each token to integrate global information and to build rich contextual representations during large-scale pretraining~\cite{cordonnier2020relationship}. 
Despite high expressive power, self-attention introduces substantial overhead: its pairwise token interactions require $\mathcal{O}(T^2)$ computation and generate heavily data-dependent memory traffic~\cite{tay2022efficient}. 
These properties align well with highly parallel digital processors such as GPUs and TPUs, but they fundamentally conflict with the computation model of emerging in-memory computing (IMC) accelerators.

IMC systems, particularly ReRAM crossbars, excel when computation can be expressed as weight-stationary vector–matrix multiplications. 
As illustrated in Fig.~\ref{fig:crossbar}, applying input voltages along wordlines and sensing accumulated currents on bitlines naturally realizes analog GEMM with minimal data movement and high energy efficiency~\cite{yang2020reram, wolters2024memory}. 
However, attention is dominated not by weight-stationary GEMM, but by \emph{activation-to-activation} multiplications such as $QK^\top$, softmax normalization, and per-token dynamic mixing. 
These operations require repeatedly reading spatially distributed activations and routing intermediate results across arrays, leading to fragmented analog operations, extensive ADC/DAC conversions, and poor crossbar utilization. 
Consequently, many IMC accelerators offload attention to digital compute units, while only mapping feed-forward layers onto ReRAM, leaving the quadratic attention bottleneck unresolved at inference time.

Meanwhile, recent analyses of transformers consistently show that attention layers exhibit substantial redundancy at inference time~\cite{bhojanapalli2021leveragingredundancy, he2024matterstransformers}. 
Across depth, attention maps tend to become progressively smoother and more compressible, and many heads can be well-approximated by low-rank or weakly varying interaction patterns. 
Based on these observations, we posit that the functional role of an attention block during inference is often closer to a stable, smooth sequence-to-sequence mapping than to a fully dynamic all-pairs interaction. 
This motivates our investigation into whether such structured behavior can be captured by sequential modules whose computation patterns naturally align with the locality and weight-stationarity properties preferred by IMC hardware.

Motivated by these observations, we introduce \textbf{FAR}, a framework that replaces every attention block in a pretrained transformer with a sequential module whose computation aligns with IMC execution, and trains them through head-level, block-wise distillation so that it faithfully replicates the original attention behavior.
All transformer components outside the replaced blocks remain frozen during distillation, ensuring that FAR functions as a drop-in architectural conversion rather than training from scratch. The resulting model thus preserves both the pretrained representation quality and the downstream transfer performance of the original transformer.

\begin{figure*}[t]
\centering
\hspace{35pt} 
\includegraphics[width=.8\linewidth]{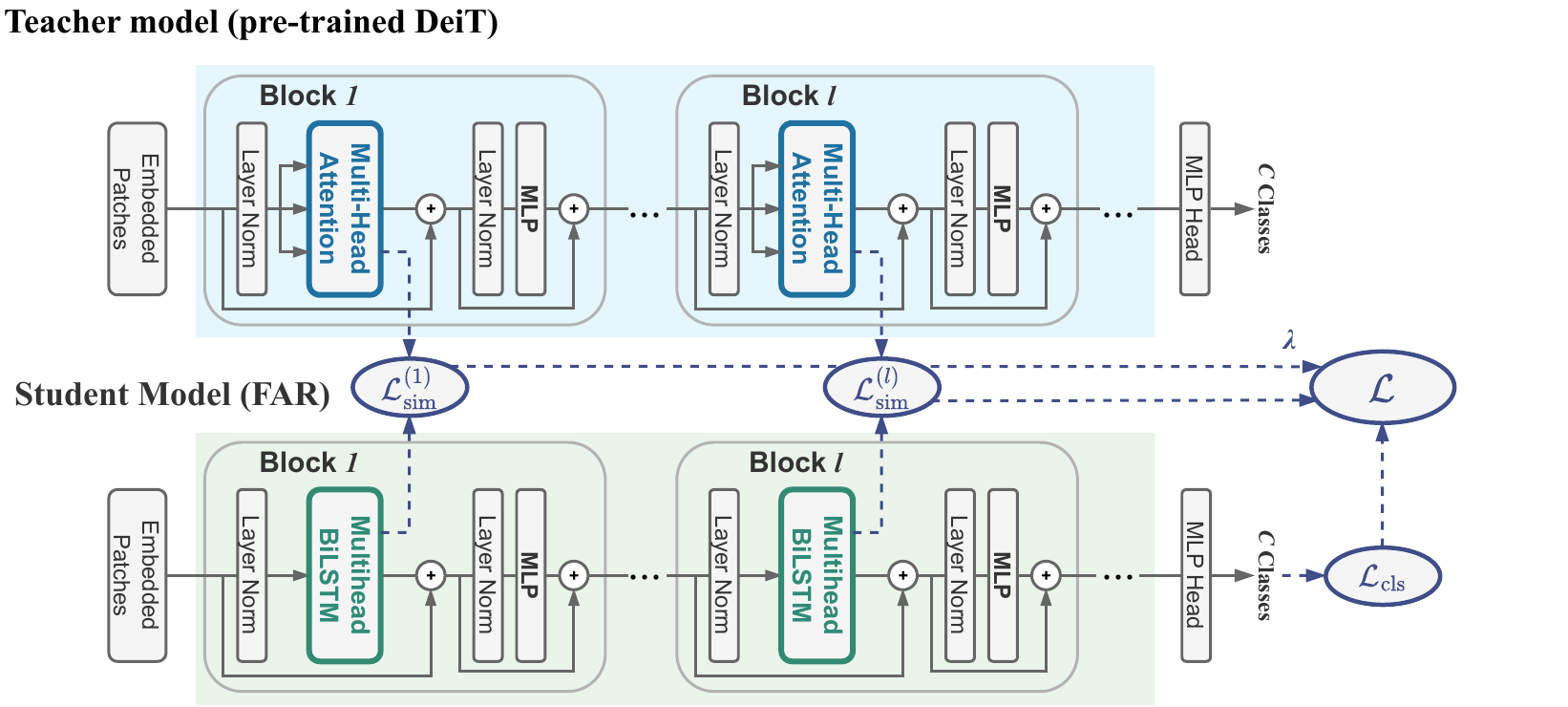}
\caption{
Block-wise replacement of attention. 
Each replaced module is supervised by a similarity loss, while a classification loss is applied at the output. 
During distillation, only the replacement blocks are updated.
}
\Description{
Block-wise replacement of attention. 
Each replaced module is supervised by a similarity loss, while a classification loss is applied at the output. 
During distillation, only the replacement blocks are updated.
}
\label{fig:loss}
\end{figure*}

We adopt an LSTM-based module as the sequential substitute because its computation pattern is inherently more compatible with IMC execution: LSTMs~\cite{hochreiter1997long} reuse a fixed set of weights across time steps, avoid activation-to-activation multiplications, and follow a strictly sequential dataflow with high weight locality. These properties map naturally onto ReRAM crossbars, in sharp contrast to attention’s non-local activation mixing and quadratic interaction pattern, which incur significant data movement on IMC hardware.
To retain the inductive bias of multi-head attention, where different heads operate on distinct subspaces and learn complementary token-mixing patterns, we design a \textbf{multi-head bidirectional LSTM} (BiLSTM) structure in which each head processes a projected slice of the embedding. This organization enables the recurrent heads to specialize in different interaction patterns, and the bidirectional recurrence provides forward and backward contextual coverage that reflects the contextual behavior of self-attention.

To further reduce the computational footprint for IMC deployment, we apply sparsity-inducing regularization~\cite{yang2020deephoyer} on the hidden dimensions of each LSTM head, enabling layer-adaptive compression that removes redundant recurrent channels while respecting LSTM gate coupling. This yields compact sequential modules whose capacity matches the functional complexity of each layer and better fits the limited array sizes and bandwidth of IMC hardware.

We validate FAR on the DeiT~\cite{touvron2021deit} vision transformer family and show that it preserves the accuracy scaling behavior of attention-based models across model sizes. FAR slightly improves performance on DeiT-Tiny and remains competitive on DeiT-Small and DeiT-Base, demonstrating that sequential substitutes can retain the representational behavior learned during large-scale pretraining even without all-pairs token interactions. FAR models also show good generalizability when finetuned to downstream classification tasks, indicating that the learned sequence-to-sequence mappings remain effective across domains. 
Beyond algorithmic quality, hardware-oriented ReRAM simulations show that FAR enables a sequential and weight-stationary execution pattern that substantially reduces memory traffic and improves inference latency on IMC accelerators. These results indicate that attention can be replaced end-to-end at inference time while maintaining practical accuracy, providing a viable and hardware-efficient alternative for deploying pretrained transformers on IMC-based systems.

\section{Related Work}
\label{sec:relatedwork}

\paragraph{Efficient Alternatives to Self-Attention.}
The quadratic cost of self-attention has motivated both attention-free and attention-efficient architectures. Attention-free models such as MLP-Mixer~\cite{tolstikhin2021mlpmixer}, RetNet~\cite{sun2023retnet}, and Mamba~\cite{gu2024mamba} replace attention with MLP-based, recurrent, or state-space token mixing, achieving linear-time complexity and improved hardware efficiency. However, these models are trained from scratch and cannot leverage pretrained transformers, limiting their applicability in scenarios that rely on large-scale pretraining. Efficient attention variants, including Linformer~\cite{wang2020linformer}, Performer~\cite{choromanski2022rethinking}, FlashAttention~\cite{dao2022flash}, and Reformer~\cite{kitaev2020reformer}, reduce attention overhead through low-rank projection, kernelization, or optimized execution. Yet these designs preserve the core all-pairs interaction pattern and dynamic memory access, which remain challenging for IMC hardware. In contrast, our work removes attention entirely during inference and replaces it with a sequential module explicitly aligned with IMC dataflows.

\paragraph{Distillation and Modular Replacement.}
Knowledge distillation~\cite{hinton2015distilling} provides a mechanism to transfer behavior from large models to compact ones. Works such as TinyBERT~\cite{jiao2020tinybert}, MobileBERT~\cite{sun2020mobilebert}, and MiniLM~\cite{wang2020minilm} introduce intermediate-layer supervision to achieve attention compression. Recent efforts further demonstrate the feasibility of distilling pretrained transformers into recurrent architectures~\cite{NEURIPS2024_mambainllama}, suggesting that attention behavior can be approximated without all-pairs computation. These approaches, however, either retain much of transformer structure or require end-to-end retraining. Our method differs by freezing the pretrained backbone and replacing each attention block through block-wise distillation, enabling functional substitution with minimal retraining while preserving compatibility with existing pretrained models.

\paragraph{Attention Acceleration on IMC Hardware.}
A number of IMC-oriented accelerator designs focus on mapping the feedforward or projection layers of transformers to crossbars while offloading attention computation to external digital processors or keeping it largely unoptimized for IMC execution~\cite{Chen2025DemonstrationOT,9580474}. Other works attempt to accelerate attention directly on analog arrays~\cite{yang2020reram,X-Former2023,li2025attar}, but still preserve its non-local activation mixing and quadratic token interaction pattern. In contrast, our approach removes attention entirely and replaces it with a sequential, weight-stationary module whose computation naturally aligns with IMC dataflows.

\section{Function-preserving Attention Replacement}
\label{sec:method}

\subsection{Replacement Strategy}
\label{sec:strategy}

Figure~\ref{fig:loss} provides an overview of our proposed FAR framework. We replace all attention modules in a pretrained transformer with multihead BiLSTM modules supervised via both the layer-wise distillation and the global task loss.
Considering a transformer composed of $L$ layers, where each layer consists of an attention block $A_l$ and a feedforward block $F_l$, connected via residual paths and layer normalization. The $l$-th layer processes input $\mathbf{x}_l \in \mathbb{R}^{T \times h}$ (where $T$ is the sequence length and $h$ is the embedding size) as:
\begin{equation}
\mathbf{y}_l = \mathbf{x}_l + A_l(\mathrm{LN}_1(\mathbf{x}_l)) \;;\quad
\mathbf{x}_{l+1} = \mathbf{y}_l + F_l(\mathrm{LN}_2(\mathbf{y}_l))
\end{equation}

The core operation in $A_l(\cdot)$ is softmax attention, which performs all-pairs activation multiplication to mix token representations. 
    This operation is precisely the component we aim to eliminate: it induces the quadratic activation interaction and non-local dataflow that the sequential substitute is designed to avoid. Our objective is to replace each attention block $A_l$ with a learnable substitute $A_l'$, such that the modified layer becomes:
\begin{equation}
\mathbf{y}_l' = \mathbf{x}_l + A_l'(\mathrm{LN}_1(\mathbf{x}_l)) \;;\quad
\mathbf{x}_{l+1}' = \mathbf{y}_l' + F_l(\mathrm{LN}_2(\mathbf{y}_l'))
\end{equation}

Replacing $A_l$ with $A_l'$ imposes three technical constraints:

\begin{itemize}
    \item \textbf{Functional equivalence:} $A_l'$ must reproduce the attention block’s output mapping, ensuring $A_l'(\mathbf{x}_l) \approx A_l(\mathbf{x}_l)$.

    \item \textbf{Plug-in compatibility:} $A_l'$ must preserve the tensor shapes and interface of $A_l$, enabling replacement without modifying any other transformer components.

    \item \textbf{Sequential execution pattern:} $A_l'$ must rely on recurrent or sequential weight reuse rather than all-pairs activation interactions, forming a linear, weight-stationary dataflow.
\end{itemize}

To enforce functional alignment, we apply layer-wise distillation. 
For each replaced layer $l$, the similarity loss is
\begin{equation}
\mathcal{L}_{\text{sim}}^{(l)}
    = \left\| 
        A_l(\mathbf{x}_l)
        - 
        A_l'(\mathbf{x}_l)
      \right\|_2^{2},
\end{equation}

where both $A_l(\mathbf{x}_l)$ and $A_l'(\mathbf{x}_l)$ denote the concatenated head-wise outputs \emph{before} the output projection, so that $\mathcal{L}_{\text{sim}}$ aligns the per-head representational structure of the original attention.

The overall training objective combines both structural and task-level supervisions:
\begin{equation}
\mathcal{L} = \lambda \cdot \sum_{l \in \mathcal{R}} \mathcal{L}_{\text{sim}}^{(l)} + \mathcal{L}_{\text{cls}},
\end{equation}
where $\mathcal{L}_{\text{cls}}$ is the cross-entropy loss for classification task output and $\lambda$ controls the weight of similarity distillation.

\paragraph*{Training protocol.} We adopt a \textit{global replacement} strategy: all attention blocks are substituted at once, avoiding representation mismatch that arises when partially replacing layers in sequence. Training proceeds in two stages:
\begin{itemize}
    \item \textbf{Distillation phase:} The substitute modules are trained through $\mathcal{L}$ while the rest of the model is frozen.
    \item \textbf{Finetuning phase:} All parameters are jointly optimized through $\mathcal{L}_{\text{cls}}$ to recover any residual accuracy drop.
\end{itemize}

\begin{figure*}[tb]
  \centering
  \includegraphics[width=.8\linewidth]{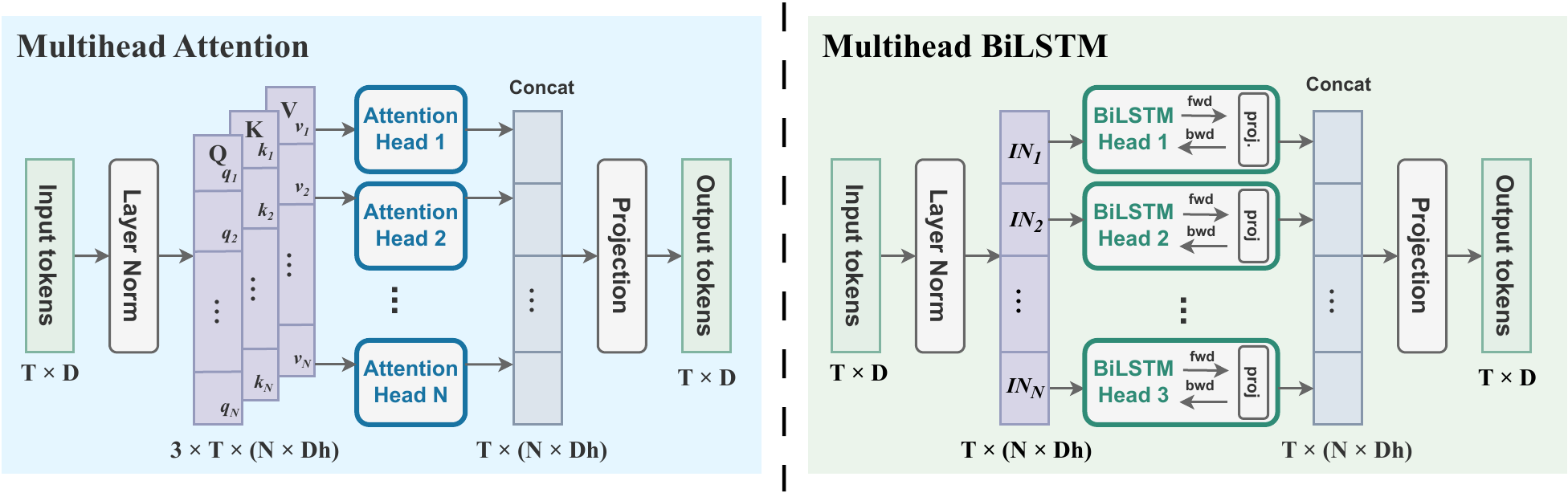}
  \caption{Multihead BiLSTM module used to replace attention. The input is first projected into $N$ subspaces and split by head. Each subspace is processed by a BiLSTM, and the outputs are concatenated and projected back to the original hidden size.}
  \Description{Multihead BiLSTM module used to replace attention. The input is first projected into $N$ subspaces and split by head. Each subspace is processed by a BiLSTM, and the outputs are concatenated and projected back to the original hidden size.}
  \label{fig:multihead}
\end{figure*}

\subsection{Multihead BiLSTM Architecture}
\label{sec:multihead-lstm}

We implement the substitute module $A_l'$ as a \textbf{multi-head bidirectional LSTM} (BiLSTM) tailored to the transformer setting and IMC execution. 
This choice introduces recurrence into each block while preserving a projection–split–recombine pattern similar to multi-head attention, and reshapes the computation into a sequential, weight-stationary dataflow that is easier to map onto IMC arrays.

The design serves two purposes. 
First, BiLSTMs model forward and backward token dependencies through localized recurrent updates, matching the sequential dataflows favored by IMC hardware. 
Second, multi-head attention scales the embedding and captures diverse token interaction patterns via separate heads. To retain this inductive bias under an LSTM-based substitute, we organize the replacement as a similar multihead structure where each head is responsible for a projected subspace of the embedding.

The overall architecture is illustrated in Figure~\ref{fig:multihead}. 
Concretely, given an input $\mathbf{x} \in \mathbb{R}^{T \times D}$, where $T$ is the sequence length and $D$ the embedding size, we first apply layer normalization and a linear projection to map it into $N$ subspaces of dimension $D_h$ ($D = N \cdot D_h$), analogous to the QKV projections in attention. Each subspace is processed by a bidirectional LSTM head to yield 
$\mathbf{H}_n = \mathrm{BiLSTM}_n(\mathbf{I}_n) \in \mathbb{R}^{T \times 2D_h}$, 
where the forward and backward hidden states are concatenated along the embedding dimension. All head outputs are then concatenated as 
$\mathbf{H} = [\mathbf{H}_1; \ldots; \mathbf{H}_N] \in \mathbb{R}^{T \times 2D}$, supervised during distillation to align the functional behavior of each head with its corresponding teacher attention head. Afterward, a linear projection is applied to restore the original dimension $\mathbf{y} = \mathrm{Proj}(\mathbf{H}) \in \mathbb{R}^{T \times D}$.

To ensure dimensional compatibility and alignment, we match the number of BiLSTM heads to the number of attention heads in the teacher model and set each head’s hidden size equal to the teacher's per-head dimension. This structural alignment allows direct replacement without modifying the surrounding architectures. 

Compared to attention, the resulting LSTM-based block eliminates all-pairs token interaction and replaces it with localized recurrence and weight reuse along the sequence, yielding linear-time inference and a computation pattern that fits naturally with IMC-oriented dataflows.

\begin{figure}[tb]
  \centering
  \includegraphics[width=0.9\linewidth]{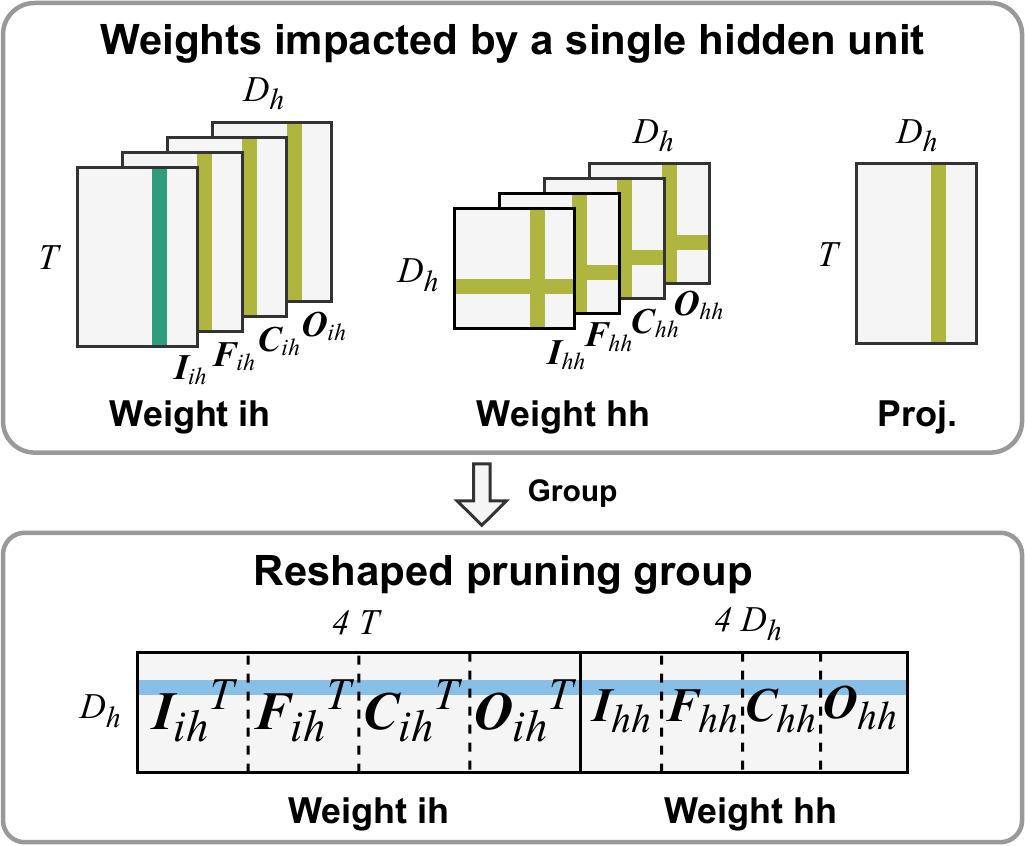} 
  \caption{Structured pruning of LSTM hidden units. Removing one unit (shaded row) consistently prunes its input–hidden weights, hidden–hidden weights, and downstream projections. Coordinated pruning across all gate matrices preserves temporal alignment.}
  \Description{Structured pruning of LSTM hidden units. Removing one unit (shaded row) consistently prunes its input–hidden weights, hidden–hidden weights, and downstream projections. Coordinated pruning across all gate matrices preserves temporal alignment.}
  \label{fig:prune}
\end{figure}

\subsection{Post-distillation Compression via Structured Pruning}
\label{sec:pruning}

\begin{table*}[t]
\centering
\caption{
Top-1 accuracy of DeiT and FAR-IMCs on ImageNet and downstream classification benchmarks. For downstream tasks, models are finetuned from pretrained FAR and DeiT models without distillation. 
}
\label{tab:accuracy}
\resizebox{0.95\textwidth}{!}{
\begin{tabular}{lccccccccc}
\toprule
\textbf{Model} & \textbf{Params (M)} & \textbf{FLOPs (G)} & \textbf{ImageNet} & \textbf{CIFAR-10} & \textbf{CIFAR-100} & \textbf{Cars} & \textbf{Flowers} & \textbf{iNat-18} & \textbf{iNat-19}\\
\midrule
DeiT-Tiny                              & 5.7 & 1.25 & 72.2 & 97.9 & 85.7 & 90.5 & 97.4 & 62.4 & 72.1 \\ 
\rowcolor{orange!20}FAR-Tiny           &7.5 & 1.45 & 73.3 & 98.1 & 85.4 & 92.3 & 97.9 & 64.7 & 71.3 \\ 
\rowcolor{orange!20}FAR-Tiny$^\dagger$ & 5.7 & 1.08 & 72.4 & 97.2 & 84.3 & 89.5 & 96.9 & 62.9 & 69.6 \\ 
\midrule
DeiT-Small                              & 22.1 & 4.60 & 79.8 & 98.5 & 87.1 & 91.7 & 98.1 & 66.8 & 74.2 \\ 
\rowcolor{orange!20}FAR-Small           & 23.9 & 4.82 & 77.8 & 98.3 & 86.8 & 92.1 & 97.9 & 67.0 & 73.9 \\ 
\rowcolor{orange!20}FAR-Small$^\dagger$ & 20.3 & 3.90 & 75.7 & 97.5 & 85.3 & 90.5 & 96.4 & 65.4 & 72.5 \\ 
\midrule
DeiT-Base                               & 86.5 & 17.56 & 81.8 & 99.1 & 90.8 & 92.1 & 98.4 & 73.2 & 77.7 \\
\rowcolor{orange!20}FAR-Base            & 83.2 & 17.31 & 79.7 & 98.2 & 87.7 & 92.3 & 97.9 & 67.0 & 75.1 \\ 
\rowcolor{orange!20}FAR-Base$^\dagger$  & 73.0 & 14.80 & 77.5 & 97.9 & 87.0 & 89.8 & 96.2 & 66.0 & 74.4  \\  
\bottomrule
\end{tabular}
}
\parbox{0.95\linewidth}{\footnotesize
$^\dagger$~Pruned model after structured compression.
}
\end{table*}

Both the pretrained transformer and our proposed LSTM-based modules introduced in Section~\ref{sec:multihead-lstm} share a unified block-wise architecture. However, prior studies reveal that redundancy varies significantly across transformer layers~\cite{yang2023global,yin2022vit}, suggesting that uniform capacity is not required at all depths.
To further improve compactness and efficient IMC mapping, we apply structured pruning to adapt the hidden dimension of each LSTM block.

\paragraph*{Extension of DeepHoyer to LSTM}
To promote structured sparsity, we extend the DeepHoyer framework~\cite{yang2020deephoyer},  which introduces a differentiable approximation to the Hoyer sparsity measure. This measure captures the sparsity of $W$ by encouraging a few large entries while suppressing the rest. DeepHoyer generalizes Hoyer measure to Group-HS for structured sparsity using group $\ell_2$ norms:
\begin{equation}
G_H(W) = \frac{ \left( \sum_{g=1}^G \| w^{(g)} \|_2 \right)^2 }{ \sum_{g=1}^G \| w^{(g)} \|_2^2 },
\end{equation}
where each $w^{(g)}$ denotes a group of weights.

While originally designed for convolutional and fully connected layers, we extend this method to bidirectional LSTM modules by targeting the hidden dimension. Because LSTMs couple all four gate matrices through shared hidden units, removing a single hidden unit must be coordinated across all gates and associated input–hidden, hidden–hidden, and projection matrices to maintain temporal and structural consistency.
Specifically, for each unidirectional LSTM block, removing a hidden unit affects all four gate matrices (input, forget, cell, and output). The structurally coupled parameters include:  
(1) the corresponding rows in the input-to-hidden matrices $W_{\text{ih}}$,  
(2) the corresponding rows and columns in the hidden-to-hidden matrices $W_{\text{hh}}$, and  
(3) the columns in the projection feeding the next layer.  
As illustrated in Figure~\ref{fig:prune}, these gate-aligned components are concatenated into a composite matrix $W^{(l)} \in \mathbb{R}^{D_h \times G}$, where each row aggregates all parameters associated with one hidden unit.

We then compute the structured regularization penalty for block $l$ across the rows of $W^{(l)}$ to obtain scores per hidden unit:
\begin{equation}
\mathcal{R}_{\text{Hoyer}}^{(l)} = \frac{ \left( \sum_{j=1}^{D_h} \| W^{(l)}[j, :] \|_2 \right)^2 }{ \sum_{j=1}^{D_h} \| W^{(l)}[j, :] \|_2^2 }.
\end{equation}
This term is added to the original loss during regularized training, guiding the network towards sparse hidden representations while maintaining structural integrity across gates.

We apply structured pruning through a three-stage pipeline:

\begin{itemize}
    \item \textbf{Regularization.} A Hoyer-based penalty is introduced to promote structured sparsity along the hidden dimension of each LSTM. This regularization can be applied either jointly with block-level distillation or as a separate retraining stage after substitute modules are initialized.
    
    \item \textbf{Pruning.} After regularized training, row-wise group norms of $W^{(l)}$ are computed and hidden units below a threshold are removed. Because each hidden unit spans all four gates, pruning is applied in a coordinated manner across all associated parameters.

    \item \textbf{Finetuning.} The pruned model is further finetuned with masked entries fixed to zero, recovering accuracy without diminishing the compression ratio.
\end{itemize}

Pruning is performed independently for each head and direction, allowing layer-adaptive sparsity. Unlike unified pruning with a fixed budget, this approach reflects the differing redundancy of layers, and its per-block adaptivity facilitates hardware-aware acceleration. 

\begin{figure*}[t]
  \centering
  \includegraphics[width=\linewidth]{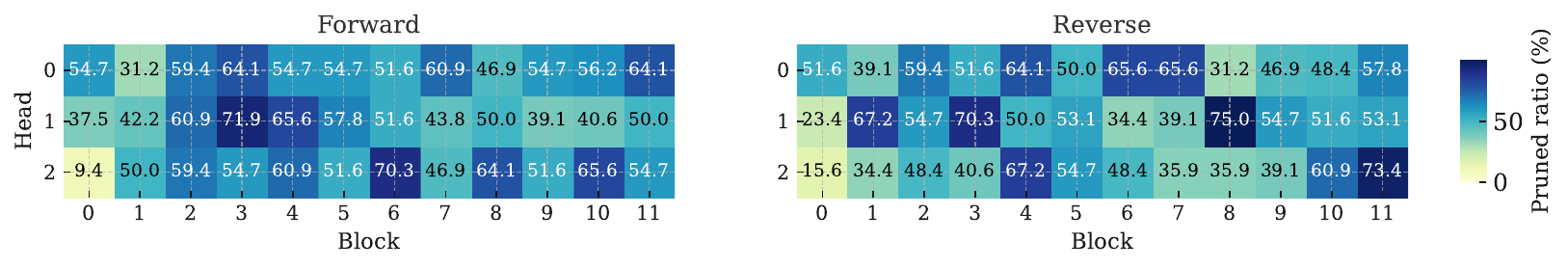}
\vspace{-20pt}
\caption{Pruning ratios across heads and directions. FAR learns to prune differently across layers and directions, revealing internal heterogeneity in representational redundancy.}
\Description{Pruning ratios across heads and directions. FAR learns to prune differently across layers and directions, revealing internal heterogeneity in representational redundancy.}
\label{fig:sparsity_pattern}
\end{figure*}

\section{Experiment results}
\label{sec:results}

\subsection{Experiment Setup}
\label{sec:setup}

\paragraph{Models and Replacement Configuration.}
We evaluate FAR on the DeiT vision transformer family (DeiT-Tiny/Small/Base)~\cite{touvron2021deit}. For each scale, all self-attention blocks are replaced with multihead BiLSTM modules of architecture in Section~\ref{sec:multihead-lstm}.
The number of heads and the per-head hidden size are matched to the teacher model to preserve dimensional compatibility.
All remaining components, including MLP blocks and patch embeddings, retain their pretrained weights and remain fixed during the distillation stage.

\paragraph{Training and Distillation Protocol.}
Training on ImageNet~\cite{deng2009imagenet} follows DeiT pipeline and augmentation settings. As described in Section~\ref{sec:strategy}, During the distillation stage, the substitute modules are updated under supervision with base learning rate $5\mathrm{e}{-4}$, AdamW optimizer, and similarity weight $\lambda = 1$. In the finetuning stage, the whole model is finetuned for 100 epochs under learning rate $5\mathrm{e}{-5}$. 
We also report Top-1 accuracy on downstream classification tasks, including CIFAR-10/100~\cite{krizhevsky2009learning}, Stanford Cars~\cite{krause20133d}, Flowers-102~\cite{nilsback2008automated}, and iNaturalist-18/19~\cite{vanhorn2018inaturalist} using
standard DeiT finetuning recipe.
Structured pruning procedure (Section~\ref{sec:pruning}) is applied to the ImageNet-trained FAR models, and the pruned checkpoints are directly finetuned on downstream tasks under the same protocol.

\paragraph{Analytical IMC Efficiency Evaluation.}
We assess hardware efficiency using an analytical IMC model.
Under a fixed crossbar configuration and device parameters derived from publicly reported ReRAM macros, we decompose each layer into VMM operations and activation movements and estimate latency and
energy using operation-level costs.
All baselines and FAR models are evaluated under identical modeling assumptions; therefore we report normalized latency and energy relative to the attention baseline.

\begin{figure}[t]
\centering
\includegraphics[width=.95\linewidth]{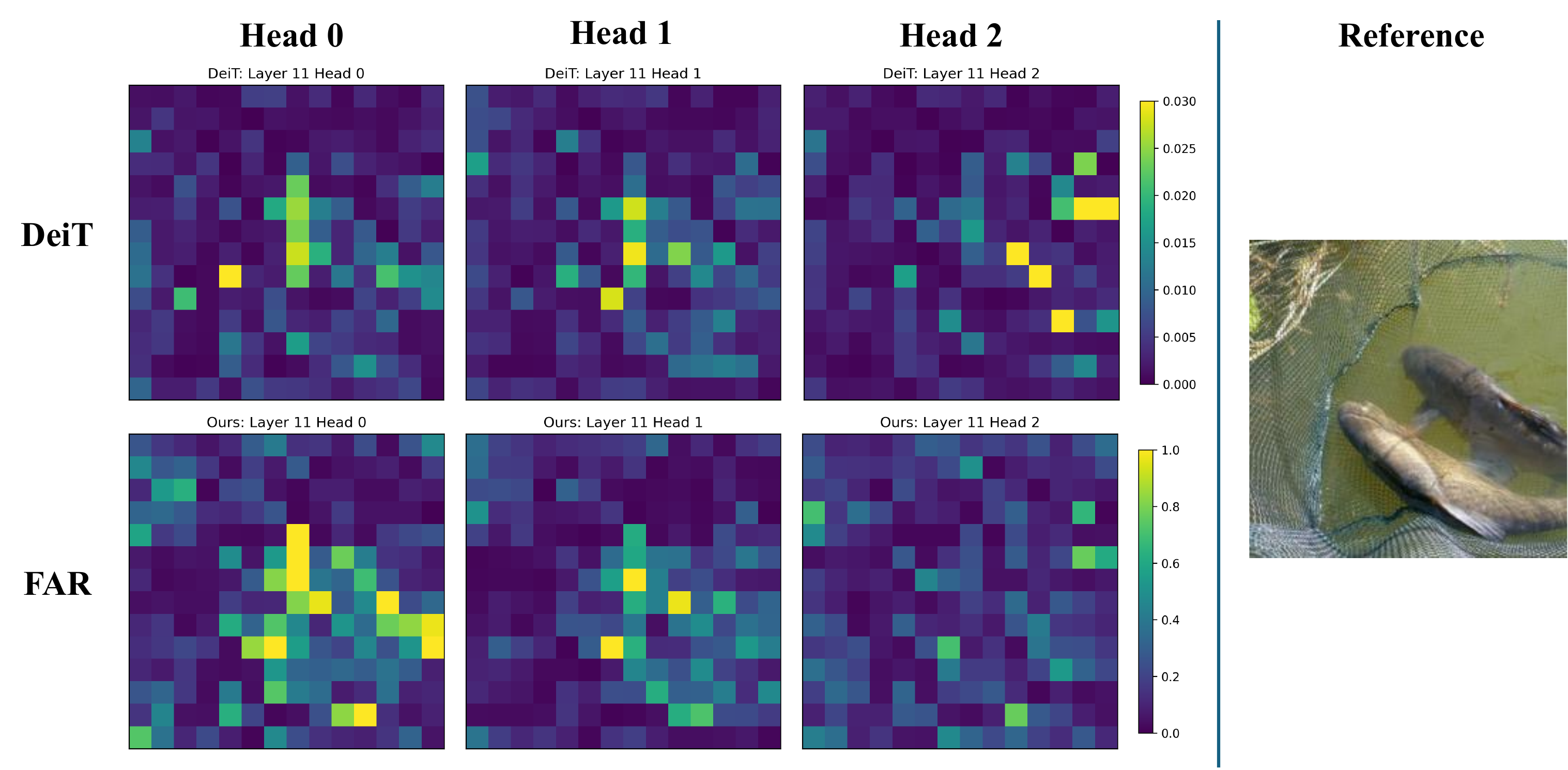}
\caption{Head-wise token interaction visualization of DeiT-Tiny and FAR-Tiny in the final transformer block. }
\Description{Head-wise token interaction visualization of DeiT-Tiny and FAR-Tiny in the final transformer block. }
\label{fig:visual}
\end{figure}

\subsection{Accuracy on Conventional Platforms}
\label{sec:accuracy}

Table~\ref{tab:accuracy} summarizes accuracy on ImageNet and downstream datasets. 

\textbf{(1) FAR preserves the functional role of attention despite removing all pairwise interactions.} 
On DeiT-Tiny, FAR exceeds the baseline accuracy by \textbf{+1.2\%}, suggesting that the sequential substitute acts as an effective inductive bias at low capacity and stabilizes token mixing where attention is over-parameterized. 
As model size increases, attention becomes more expressive and the relative gap between FAR and the teacher gradually widens, yet remains modest ($\sim$1–2\%) on both ImageNet and all downstream datasets. 
Given that attention contributes the majority of the transformer's expressivity, such a small gap indicates that the token-to-token mixing performed by attention is largely reproducible through sequential recurrence once head-wise representations are properly aligned. 
This stability confirms that FAR continues to scale with depth and width, and that its deviation from attention represents a bounded functional difference, highlighting its potential as a structurally simplified and efficient alternative to attention.

\textbf{(2) FAR retains the transferability of attention-based representations.}
On downstream benchmarks, FAR achieves accuracy comparable to the DeiT teacher across nearly all datasets, indicating that removing pairwise attention does not hinder the reuse of pretrained semantic structure. The ability to maintain task-level generalization after full replacement demonstrates that the sequential substitute modules do not merely regress to local pattern modeling, but effectively learn a transferable approximation of attention’s representational function through distillation.

\textbf{(3) FAR remains robust under structured compression.}
Applying structured pruning to FAR models leads to only an additional $\sim$1--2\% accuracy drop, even though the average hidden dimension across BiLSTM blocks is reduced by 40--60\%. As shown in Figure~\ref{fig:sparsity_pattern}, pruning per block yields a non-uniform sparsity pattern: higher retention rates are observed in middle layers while more weights are pruned towards the two ends. This pattern reflects the varying representational redundancy along depth and is consistent with prior observations on transformer compression across layers~\cite{yang2023global}. The pruned models follow the same accuracy trend as the unpruned counterparts, indicating that DeepHoyer-based structured regularization removes redundant hidden units without breaking the functional alignment established during distillation. This robustness under substantial, depth-adaptive compression provides strong evidence that FAR substitutes retain flexible capacity allocation and are suitable for IMC deployment, where smaller hidden dimensions translate directly to better array utilization and lower latency.

\textbf{(4) FAR preserves head-specific interaction patterns.}
Figure~\ref{fig:visual} visualizes the head-wise token interaction maps of the final block for DeiT-Tiny and FAR-Tiny. 
Different heads in DeiT emphasize distinct subregions of the target object, reflecting their complementary semantic roles rather than redundant attention. 
The corresponding FAR heads display highly similar activation structure: each concentrates on the foreground object region while maintaining the original head-to-head specialization pattern. 
This suggests that the sequential substitutes not only retain global classification capability but also inherit the distributed representational roles historically assigned to individual heads.

Overall, the accuracy results show that full attention removal is feasible: FAR preserves the majority of the pretrained transformer's capability, scales reliably across model sizes, and remains resilient under structured pruning while producing architectures substantially more compatible with IMC execution.

\begin{table}[t]
\centering
\captionsetup{width=0.75\linewidth}
\caption{Estimated latency and energy per DeiT-Tiny block (Attention+FFN) on different backends, normalized to that of FAR-Tiny on ReRAM IMC. Only crossbar VMM reads/writes inside each block are counted.}
\label{tab:efficiency}
\begin{tabular}{lcc}
\toprule
Model / Backend      & Latency        & Energy \\ \midrule
\rowcolor{orange!20}FAR (ReRAM IMC) & $1\times$   & $1\times$  \\
DeiT (ReRAM IMC)& $\approx 18\times$    & $\approx 3\times$    \\
DeiT (GPU)      & $\approx 400\times$   & $\approx 150\times$  \\
\bottomrule
\end{tabular}
\end{table}

\subsection{Latency and Energy Estimation}
\label{sec:efficiency}

To understand the hardware benefits brought by replacing attention with BiLSTM, we estimate the per-block latency and energy of DeiT-Tiny and FAR-Tiny under a ReRAM-based in-memory computing (IMC) backend, and relate these estimates to a conventional GPU execution. 
We focus only on the arithmetic inside each Transformer block and count the number of vector–matrix multiply (VMM) reads and dynamic writes on the crossbar; peripheral circuitry such as DAC/ADC, routing, layer normalization, element-wise activations, and global embedding/classifier layers are excluded so that the comparison isolates the cost of token-mixing modules. 
This choice matches our algorithmic change: FAR only modifies the token-mixing blocks (attention $\rightarrow$ LSTM), while other parts of the network are shared and would contribute similarly on both backends.

For the IMC-side model, we adopt array and device parameters from existing ReRAM NPUs and surveys~\cite{survey1, wolters2024memory} and abstract a row-wise VMM read and a row-wise write as the basic crossbar operations. Device-level measurements consistently report that programming (write) is much more expensive than reading, with roughly one to two orders-of-magnitude longer latency and about an order-of-magnitude higher energy~\cite{Overcoming, NVSim, wolters2024memory}. 
We therefore fix, in all our estimates, the write latency to be $100\times$ the read latency and the write energy to be $10\times$ the read energy, and express the block cost purely in terms of how many read rows and write rows each mapping triggers. 
On the attention side, we instantiate DeiT-Tiny on IMC by following the optimized dataflow of ReTransformer~\cite{yang2020reram}, which restructures multi-head self-attention to cache the input feature matrix $X$ as $X^\top$ on the array and reuse it across queries, removing most repeated writes to $K^\top$ and $V^\top$ and leaving one dynamic write of $X^\top$ plus several VMM reads per block. 
On the LSTM side, we instantiate FAR-Tiny via an ERA-LSTM style tiled mapping~\cite{ERA-LSTM}, where all gates of a head share the same crossbar tiles and are evaluated across time steps with fixed weights, so the BiLSTM token mixer only incurs VMM reads and never programs $X^\top$-like intermediates. 
To obtain a GPU reference, we use reported ReRAM–GPU comparisons from prior PIM accelerators. The ReTransformer chip claims $23.2\times$ higher computing efficiency and $1086\times$ lower power than a GPU implementation of attention, while the memristor SoC in~\cite{tetra} reports a $49\times$ energy-efficiency gain (TOPS/W) over an NVIDIA A100. Taking these as indicative ranges, we conservatively assume that mapping DeiT-Tiny to a ReRAM IMC backend reduces energy by about $50\times$ and latency by about $20\times$ compared to a GPU for the same block-level workload. 

The resulting normalized block-level latency and energy are summarized in Tab.~\ref{tab:efficiency}. Taking DeiT-Tiny as replacing baseline, we estimate the relative ratio of different configurations compared to FAR on ReRAM IMC. Under the above assumptions, DeiT on ReRAM IMC is about one order of magnitude slower and a few times more energy-consuming than FAR, and a DeiT on a GPU is roughly two to three orders of magnitude worse than FAR (IMC) in both latency and energy. This highlights that replacing attention with IMC-friendly sequential token mixers substantially improves the latency–energy profile of DeiT-like vision Transformers on ReRAM accelerators.

Note that softmax is excluded from the estimation. Measurements on GPUs and digital accelerators show that softmax can account for about $40\%$–$60\%$ of attention runtime~\cite{STAR,Topkima}, and several ReRAM-attention designs further identify softmax as a major bottleneck on IMC because it requires repeated in-memory compare/select operations and peripheral lookup tables~\cite{yang2020reram,Softermax}. The exact cost, however, depends strongly on circuit-level choices and current ReRAM chips do not provide a stable device model for softmax. In our analytic model we therefore \emph{exclude} softmax and only count crossbar VMM reads/writes. Since DeiT applies softmax in every attention block whereas FAR only applies element-wise Sigmoid in LSTM gates, ignoring softmax makes all our numbers conservative for FAR and any realistic softmax implementation would further widen the gap between attention-based DeiT and sequential FAR on IMC.

\section{Conclusion}

We presented FAR, a function-preserving attention replacement framework that substitutes every attention block in a pretrained transformer with multi-head BiLSTM modules trained via layer-wise distillation. By aligning each substitute head with its teacher counterpart and applying structured pruning on the hidden dimensions, FAR preserves the accuracy scaling and transferability of DeiT across ImageNet and multiple downstream benchmarks while substantially reducing the effective model capacity. ReRAM-based IMC simulations further show that the resulting sequential, weight-stationary dataflow lowers memory traffic and improves end-to-end latency and energy compared to attention-based baselines, highlighting the advantage of replacing all-pairs token interactions with hardware-aligned recurrence. Taken together, these results indicate that transformer inference can be restructured around IMC-friendly sequential modules without retraining from scratch, providing a practical path toward deploying large pretrained models on emerging memory-centric accelerators.

\begin{acks}
The University of Arizona team thanks the supports of research collaboration grant and computation resources from TetraMem, Inc to this work. This work was based in part upon High Performance Computing (HPC) resources supported by the University of Arizona TRIF, UITS, and Research, Innovation, and Impact (RII) and maintained by the UArizona Research Technologies department.
\end{acks}

\bibliographystyle{ACM-Reference-Format}
\bibliography{sample-base}

\end{document}